\documentclass[conference]{IEEEtran}
\IEEEoverridecommandlockouts
\usepackage{hyperref}
\usepackage{multirow}
\usepackage{tabularx} 
\usepackage{array}
\usepackage{booktabs}
\usepackage{tabularray}
\usepackage{amsmath,amssymb,amsfonts}
\usepackage{algorithmic}
\usepackage{graphicx}
\usepackage{textcomp}
\usepackage{xcolor}
\usepackage{etoolbox}
\preto\tabular{\setcounter{magicrownumbers}{0}}
\newcounter{magicrownumbers}

\usepackage{subcaption}

\def\BibTeX{{\rm B\kern-.05em{\sc i\kern-.025em b}\kern-.08em
    T\kern-.1667em\lower.7ex\hbox{E}\kern-.125emX}}
\begin{document}

\title{Location based Probabilistic Load Forecasting of EV Charging Sites: Deep Transfer Learning with Multi-Quantile Temporal Convolutional Network}


\author{
\IEEEauthorblockN{\fontsize{6}{10}\selectfont Mohammad Wazed Ali}
\IEEEauthorblockA{\fontsize{6}{10}\selectfont \textit{Intelligent Embedded Systems (IES)}, \textit{University of Kassel}, \\ \textit{Kassel, Germany}}
\and
\IEEEauthorblockN{\fontsize{6}{10}\selectfont Asif bin Mustafa}
\IEEEauthorblockA{\fontsize{6}{10}\selectfont \textit{School of CIT}, \textit{Technical University of Munich}, \\ \textit{Munich, Germany}}
\and
\IEEEauthorblockN{\fontsize{6}{10}\selectfont Md. Aukerul Moin Shuvo}
\IEEEauthorblockA{\fontsize{6}{10}\selectfont \textit{Dept. of Computer Science and Engineering}, \\ \textit{Rajshahi University of Engg. \& Technology}, \\ \textit{Rajshahi, Bangladesh}}
\and
\IEEEauthorblockN{\fontsize{6}{10}\selectfont Bernhard Sick}
\IEEEauthorblockA{\fontsize{6}{10}\selectfont \textit{Intelligent Embedded Systems (IES)}, \textit{University of Kassel}, \\ \textit{Kassel, Germany}}
}

\maketitle
\begin{abstract}
Electrification of vehicles is a potential way of reducing fossil fuel usage and thus lessening environmental pollution. Electric Vehicles (EVs) of various types for different transport modes (including air, water, and land) are evolving. Moreover, different EV user groups (commuters, commercial or domestic users, drivers) may use different charging infrastructures (public, private, home, and workplace) at various times. Therefore, usage patterns and energy demand are very stochastic. Characterizing and forecasting the charging demand of these diverse EV usage profiles is essential in preventing power outages. Previously developed data-driven load models are limited to specific use cases and locations. None of these models are simultaneously adaptive enough to transfer knowledge of day-ahead forecasting among EV charging sites of diverse locations, trained with limited data, and cost-effective. This article presents a location-based load forecasting of EV charging sites using a deep Multi-Quantile Temporal Convolutional Network (MQ-TCN) to overcome the limitations of earlier models. We conducted our experiments on data from four charging sites, namely Caltech, JPL, Office-1, and NREL, which have diverse EV user types like students, full-time and part-time employees, random visitors, etc. With a Prediction Interval Coverage Probability (PICP) score of 93.62\%, our proposed deep MQ-TCN model exhibited a remarkable 28.93\% improvement over the XGBoost model for a day-ahead load forecasting at the JPL charging site. By transferring knowledge with the inductive Transfer Learning (TL) approach, the MQ-TCN model achieved a 96.88\% PICP score for the load forecasting task at the NRELsite using only two weeks of data.
\end{abstract}

\begin{figure}[t!]
  \centering
  \includegraphics[width=0.4\textwidth, height=0.32\textheight]{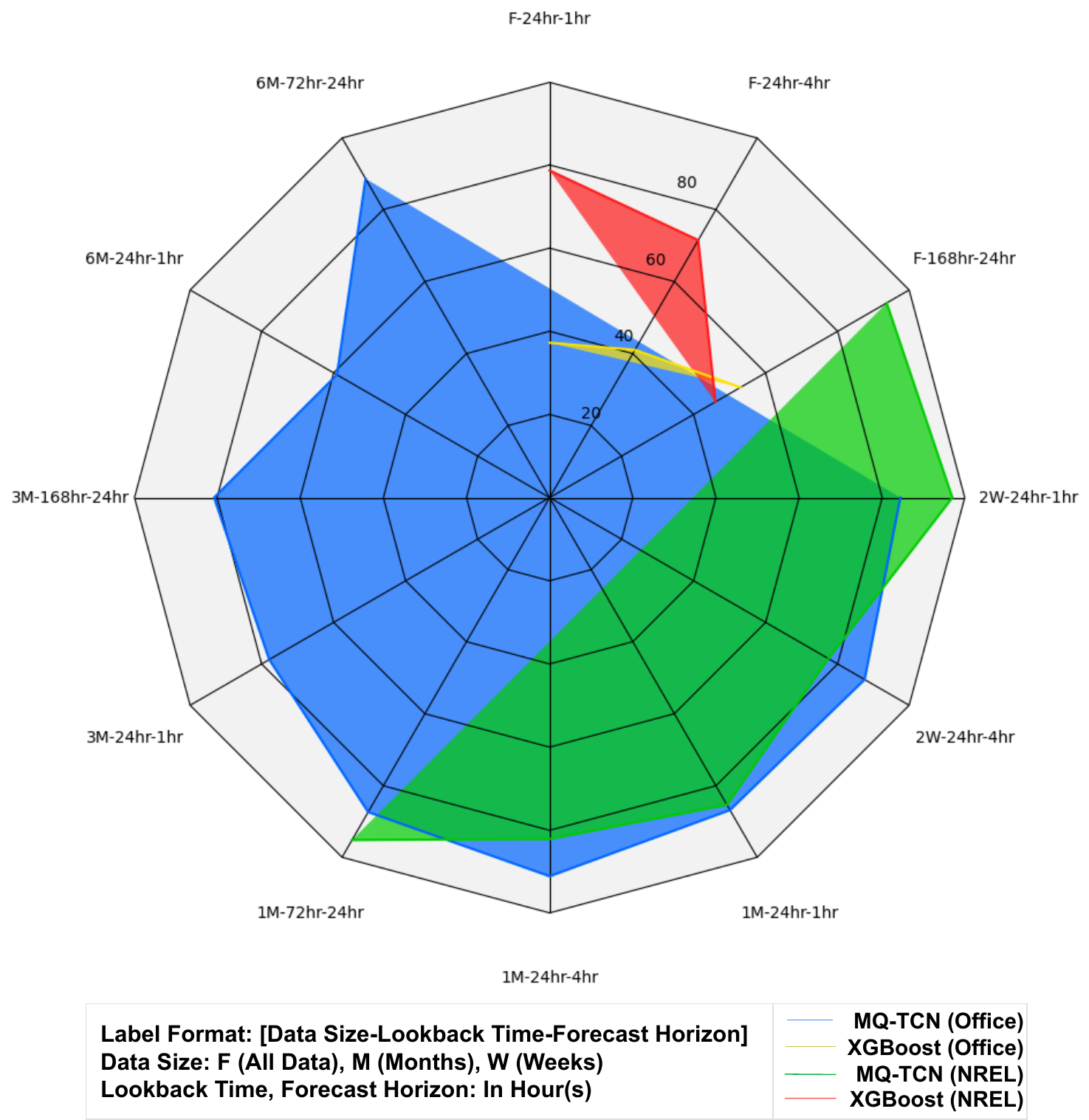}
  \caption{Load Forecasting at JPL and NREL Electric Vehicle Charging Sites using Deep MQ-TCN model with Inductive Transfer Learning.}
  \label{fig:radar_plot_office_nrel}
\end{figure}

\section{Introduction}
\label{sec:introduction}
To meet the Paris Agreement demand adopted in 2015, electric vehicles (EVs), combined with a renewable energy mix, are a great candidate to replace fossil fuel-based internal combustion engines and reduce global warming, greenhouse gas emissions, and climate change \cite{Banerji2021IntegratingRE}\cite{Singh2013CorrigendumTH}. In recent years, electric car sales have experienced exponential growth, with around 14 million new electric cars sold alone in 2023 and expected to reach 17 million in 2024 \cite{IEA2024}. 
However, this tremendous eagerness to transform the transport sector with EVs and increase the total share of the grid’s renewable energy mix brings many unique challenges, such as power outages due to excessive energy demand. Very often, there is no fixed charging behavior, and any energy demand pattern exists in advance for EVs, which may lead to uncontrolled charging and grid instability \cite{5356176}. In addition, integrated renewable energy sources depend highly on weather, location, and other climate-related factors \cite{ali2022deep}. 

Therefore, stakeholders like EV charging station operators, energy traders, and energy producers need to know the aggregated energy demand of a charging site in advance at a nearby time or for special occasions such as public holidays \cite{9069450}. Understanding EV users' aggregated energy demand at nearby time points (1-hour, 4-hour, 12-hour, 24-hour forecast horizon) of a particular charging site will give stakeholders
a significant advantage in incentivizing pricing strategies or charging EVs intelligently \cite{vuelvas2020energy}.

Many past research works based on statistical or machine learning-based specialized load models failed due to the lack of adaptability and inability to model complex interactions among heterogeneous temporal features and irregular energy usage patterns \cite{khoo2014statistical}\cite{Frendo2020ImprovingSC}. Some models are very sensitive to outliers \cite{liang2015kernel}, and some can not handle large datasets well \cite{Xydas2013ForecastingEV}\cite{majidpour2016forecasting}. In recent years, researchers have started to use deep learning (DL) models. However, much of this DL-based research was completed earlier with a large amount of data and high computational resources \cite{b12}, point forecasts were performed only without uncertainty estimates, and knowledge transfer did not occur  \cite{b1}\cite{b11}. In addition, many of these models consider only numerical and temporal features and ignore informative high cardinal categorical features \cite{10454241}. Recently, some research ideas have been discussed and analyzed to overcome the limitations of earlier models without in-depth experimental evaluations \cite{ali2022deep}\cite{9935088}. In our research, we not only analyzed but also focused on in-depth experimental evaluation of our ideas to overcome the following challenges:\\
\textbf{Firstly,} to collect an abundant amount of data in the electromobility  (e-mobility) domain is not easy as it contains sensitive personal information and due to its economic value \cite{Garca2014DataCA}. \textbf{Secondly,} training an individual DL-based load model from scratch for geographically separated charging sites is costly and time-consuming. \textbf{Thirdly,} building an efficient DL model with abundant data requires massive computation power and thus generates high temperatures, which oppose the objectives of the Paris Agreement \cite{Gielen2019TheRO}.\\
Therefore, this research considers inductive Transfer Learning (TL) from a knowledge-transferring perspective among geographically separated charging sites and ensures self-adaptive properties in the models. Rather than developing DL models for point forecasting, we consider a quantile regression-oriented probabilistic approach to capture uncertainty in combination with Temporal Convolutional Network (TCN) \cite{Xu2023QuantileRB}. In this research, we proposed Multi-Quantile TCN (MQ-TCN), an architecture described briefly in Section \ref{sec:relatedWork} novel in the e-mobility domain for load forecasting of charging sites incorporating quantile regression techniques and TCN capabilities. To evaluate MQ-TCN architecture using real-world datasets for load forecasting tasks, we are motivated to explore the following research questions in this research:
 \begin{enumerate}
     \item How to develop a multivariate multi-step load forecasting model of charging stations using a deep MQ-TCN for the e-mobility domain?
     \item How can we effectively transfer day-ahead load forecasting knowledge using a deep inductive TL approach among geographically separated EV charging sites with low computational resources?
     \item How can we handle uncertainty? 
     \item How does the performance of day-ahead load forecasting in TL settings change with lookback period, forecast horizon, and training data size?
 \end{enumerate}

This research used historical charging session data from four real-world EV charging sites: Caltech, JPL, Office-1, and NREL \cite{Neuman2021WorkplaceCD}. Here, the Caltech, JPL, and Office-1 data is obtained from the primary dataset, ' ACN' \cite{lee_acndata_2019}. Experimental evaluation of these datasets using proposed MQ-TCN architecture in comparison to the XGBoost \cite{chen2015xgboost} and Deep Auto-Regressive recurrent networks (DeepAR) \cite{Flunkert2017DeepARPF} models leads to the following contributions:

\begin{itemize}

\item \textbf{We are the first} to propose utilizing the inductive TL approach in our knowledge using a multi-step quantile regression method with deep learning architecture like multivariate TCN to forecast day-ahead load of geographically separated EV charging sites under extremely limited computational resources (data, computing power, etc.) for the e-mobility domain.

\item \textbf{Without the TL approach}, our proposed MQ-TCN architecture achieved an impressive  Prediction Interval Coverage Probability (PICP) score of 93.62\% as shown in Figure \ref{fig:radar_plot_office_nrel}, and this resulted in an improvement of around 28.93\% for one day-ahead of load forecasting than usual XGBoost model at the JPL site in the same settings. 
\item \textbf{With the inductive TL approach,} with the pretrained MQ-TCN model, using only 2 weeks of data, it reached a PICP score of 96.88\% as illustrated in Figure \ref{fig:radar_plot_office_nrel} for hourly forecasting at NREL site. This resulted in an improvement of 18.23\% compared to the XGBoost model trained on 6 months of data without TL. In similar TL settings at the Office-1 site, MQ-TCN achieved 91.04\% for 4 hours-ahead forecasting and 87.30\% for one day-ahead forecasting task using only1 month of data.
\item \textbf{To capture the uncertainty} of the model's generated load forecasts of EV charging sites, our utilized PICP and Winkler Score (WS) metrics efficiently estimate uncertainties jointly by quantifying accurate forecasting within efficient prediction intervals.

\item \textbf{By investigating the impact} of data size, lookback period, and forecasting horizon time via experimental evaluation. It has been observed that transferring knowledge with an inductive TL approach significantly reduces learnable parameter size without losing forecasting accuracy.  While applying the inductive TL approach from JPL to the Office-1 site with the same settings and using the proposed MQ-TCN model, a 72\% reduction in learnable parameter size is observed for data sizes of 2 weeks compared to 1 month. However, this reduction only compensates for 3.4\% of the PICP score. Moreover, while varying the lookback period and the length of the forecast horizon, no explicit observation has been found related to the forecasting accuracy.\\
 \end{itemize}

In this article, Section \ref{sec:relatedWork} discusses related research in e-mobility. Section \ref{sec:problemDefinition} 
 formulates problem definition, and Section \ref{sec:methodology} gives an overview of the proposed MQ-TCN model and discusses TL settings applied in this research. Section \ref{sec:experimentalEvaluation} presents details of the experimental settings, evaluation measures with results, and analysis. Section  \ref{sec:conclusion} concludes this research article and describes future work.

\section{Related Work}
\label{sec:relatedWork}

E-mobility has become a prominent field in recent years, and forecasting the energy demand of charging sites has become an essential area of research. In this paper \cite{b1}, EV charging demand forecasting models were developed for a day-ahead horizon and 15-minute resolution. The models incorporated calendar and weather data, and the LSTM model was used. The authors of \cite{b2} evaluated the use of deep neural networks (DNNs) and tree-based machine learning models for forecasting energy demand at public EV charging stations and conclude that tree-based models outperform DNNs in accuracy and error metrics. The authors of  \cite{b2} also studied how traffic, crowd distribution, weather, and charging station distribution affect charging demand \cite{b3}. In the deep learning model family, transformers perform better in e-mobility energy demand forecasting \cite{b12}. Probabilistic forecasting is an exciting research field that predicts the charging demand for e-mobility. Machine Theory of Mind Based Quantile Forecast Network (MBQFN) framework has proven to perform better than existing DeepAR, ARX-GARCH, DLQR, Persistence, and T-CKDE \cite{b13}. In an investigation of \cite{b14}, Bayesian deep learning and quantile regression methods enhanced load forecasting accuracy in EV charging stations. This approach outperforms traditional methods in dealing with uncertainties. Numerous research were completed earlier to apply to electricity load forecasting using convolutional neural networks (CNNs) \cite{b5}, \cite{b6}, \cite{b4}. Recently, some research has been conducted on the transfer of knowledge in e-mobility, where data is limited in the target domain. The results indicated that using CNN-BiLSTM has improved accuracy in predicting EV charging demand and network voltage profiles while reducing computational requirements for power grid operations, but it is unable to handle uncertainty \cite{b11}. In \cite{b30}, XGBoost was used alongside other algorithms on the ACN dataset for both classification and regression tasks with good scores at AUC, F1 Score, R², and RMSE metrics. It lacked adaptability and could not handle uncertainty. In \cite{b31}, XGBoost predicted single-step charging demand forecasting at public stations.

\section{Problem definition and formulation}
\label{sec:problemDefinition}
To formulate multi-step load forecasting task formally, let's denote a heterogeneous multivariate time series dataset of $d(d\geq 1)$ number of non-predictive variables with length $T$ as $X=\left \{ x_{t} \right \}^{T}_{t=1}=\left \{ x_{1},x_{2},....,x_{T} \right \}^{T}\in \mathbb{R}^{T\times d}$ and target variable as $Y=\left \{ y_{t} \right \}^{T}_{t=1}=\left \{ y_{1},y_{2},....,y_{T} \right \}^{T}\in \mathbb {R}^{T\times 1}$.  Here, $ x_{t}\in \mathbb{R}^{d}\left ( 1\leq t\leq T \right )$ is a vector with $d$ non-predictive variable at time step $t$ whereas $ y_{t}\in \mathbb{R}\left ( 1\leq t\leq T \right )$ symbolizes the target variable at the same time step  $t$. Based on the task definition and forecasting horizon $\delta   \left ( \delta   \geq 1 \right )$, an estimate of the target variable is the output of the load prediction model after $T+\delta$ denotes as  $ \left \{ \hat{y}_{t} \right \}^{T+\delta }_{t=T+1}=\left \{ \hat{y}_{T+1},\hat{y}_{T+2},....,\hat{y}_{T+\delta} \right \}$. In this article, we consider a deep neural network variant as a non-linear mapping function $F\left ( \bullet  \right )$ to forecast future value  $\left \{\hat{y}_{t} \right \}^{T+\delta}_{t=T+1}$ with $\delta$ time steps ahead from non-predictive input \\$\left \{ x_{t} \right \}^{T}_{t=1}$, and target series $\left \{ y_{t} \right \}^{T}_{t=1}$  and represent as  $ \left \{ \hat{y}_{t} \right \}^{T+\delta }_{t=T+1}=F\left ( \bullet  \right ) = F\left ( \left \{ x_{t} \right \}^{T}_{t=1}, \left \{ y_{t} \right \}^{T}_{t=1}\right )$. 

This research article also focuses on transferring knowledge of load forecasting tasks among EV charging sites with similar characteristics using the deep TL approach. To transfer knowledge from source (S) tasks $ \left \{ \mathcal{T}_{sd_{m}} \right \}_{m=1}^{m=M} $  in where $M\in N^{+}$ and source domain $\left \{ D_{s_{m}},\mathcal{T}_{sd_{m}} \right \}$ to a target (T) task  $\mathcal{T}_{td_{m}} $, an objective function $f_{td}\left ( . \right )$ needs to be learned with limited training samples $\left \{ x_{i_{td}},y_{i_{td}} \right \}$ in target domain. In our research, both source and target tasks $\left (  \mathcal{T}_{sd_{m}}, \mathcal{T}_{td_{m}}  \right )$ in TL settings are represented as multi-step load forecasting with the deep neural network as non-linear objective function $\left (f_{td}\left ( . \right )  \right )$ to transfer knowledge from source EV sites having abundant resources (data, computing power etc.) to resource constraint EV charging sites.
\section{Methodology}
\label{sec:methodology}
In this section, we define the related terms, methods and then describe our proposed architecture.

\subsection{
Multivariate MQ-TCN for Multi-Step Forecasting}

\begin{figure*}[htbp]
 \centering
 \resizebox{\linewidth}{!}{\includegraphics[width=0.8\textwidth, height = 1.5in]{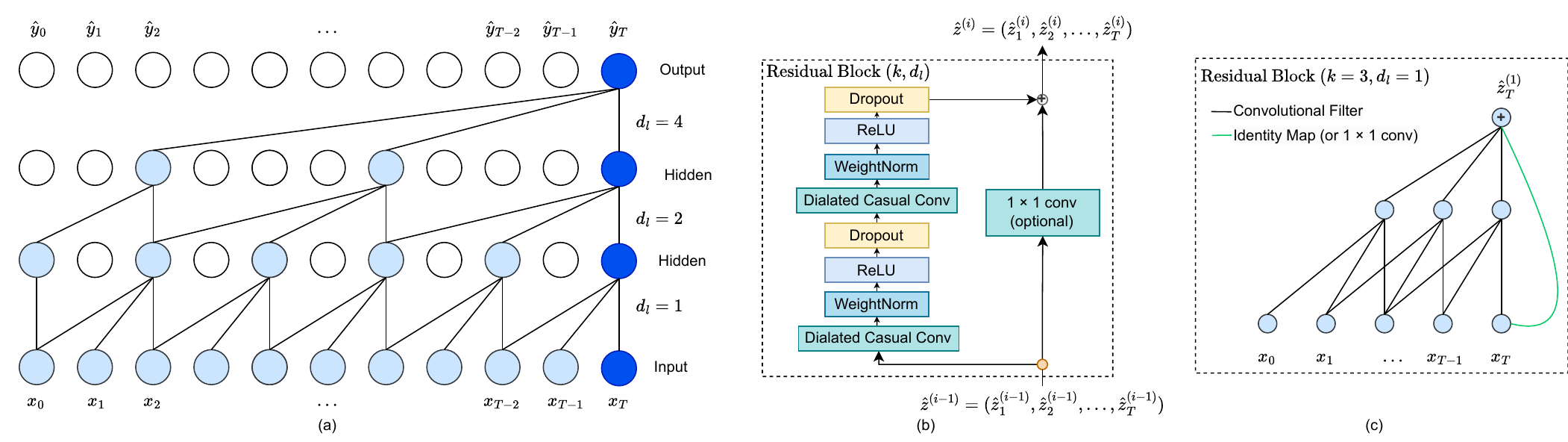}}
 \caption{Components of a TCN (a) A dilated causal convolution applied convolutional filters with dilation factors \( d_l \) = 1, 2, and 4, and a filter size of k = 3; (b) In a TCN residual block, if the input and output have different sizes, a 1 × 1 convolution is included to adjust for the difference; (c) Illustration of Residual Connection within a TCN \cite{b24}.}
 \label{fig:tcn}
\end{figure*}


TCN is a variant of CNNs that uses parallel computations consisting of three main components: causal convolution, dilated convolution, and a residual connection. One of the most important features of this system is its ability to process inputs of any length and produce outputs of the same length. This is achieved by using a 1D fully convolutional network (FCN) structure where each hidden layer matches the input layer in length. Also, zero-padding is added to maintain consistency across layers. Causal convolutions ensure that the model only focuses on historical data to prevent data leakage in the future \cite{b24}. Traditional causal convolutions have a limited reach that increases linearly with network depth, which is insufficient for long sequences or historical data. To address this limitation, TCN uses dilated causal convolutions that skip inputs at defined intervals, thus significantly expanding the receptive field's size \cite{b26}.

For a sequence input \( x_{1},x_{2},....,x_{T} \in \mathbb{R} \) and a filter \( f_{filter}: \{0,1,...,k - 1\} \), the dilated convolution function \( G \) on element \( s \) of the sequence is expressed as:
\begin{equation}
G(s) = (x *_{d_l} f_{filter})(s) = \sum_{i=0}^{k-1} f_{filter}(i) \cdot x_{s-d_l \cdot i}
\end{equation}
where \( k \) denotes the dimension of the filter, \( d_l \) is the dilation factor which determines the interval at which the input values are sampled, \( *_{d_l} \) represents the dilated convolution operator, and \( x_{s-d_l \cdot i} \) indicates a backward step through the sequence indexed from \( s \).


The receptive capacity of a TCN is inherently linked to the network's depth, represented by \( n \), and the dilation factor, denoted as \( d_l \). It is crucial to stabilize the TCN, particularly when deepening it, to overcome common issues of deep networks such as performance degradation and vanishing gradients. In this solution, a residual module is used, leveraging residual connections to enhance and simplify the network training process \cite{b25}. Such a module and its connection are illustrated in Figures 2b and 2c. To approximate a given function \( H(x) \), we can reformulate it using another function \( G(x) \) defined as \( G(x) := H(x) - x \). This allows us to recast the original learning target as \( G(x) + x \), based on the concept of residual connections. This adjustment simplifies the optimization process compared to optimizing the unaltered function \( H(x) \) directly. The outcomes are then routed back as inputs to the block, producing an output \( o \) as illustrated by:
\begin{equation}
o = \text{Activation}(x + G(x))
\end{equation}


This mechanism permits the network layers to adjust the identity mapping incrementally rather than relearning the entire transformation from scratch. This is substantially beneficial for deep neural networks' depth-wise learning capacity.

\begin{figure}[htbp]
 \centering
 \resizebox{\linewidth}{!}{\includegraphics[width=0.8\textwidth]{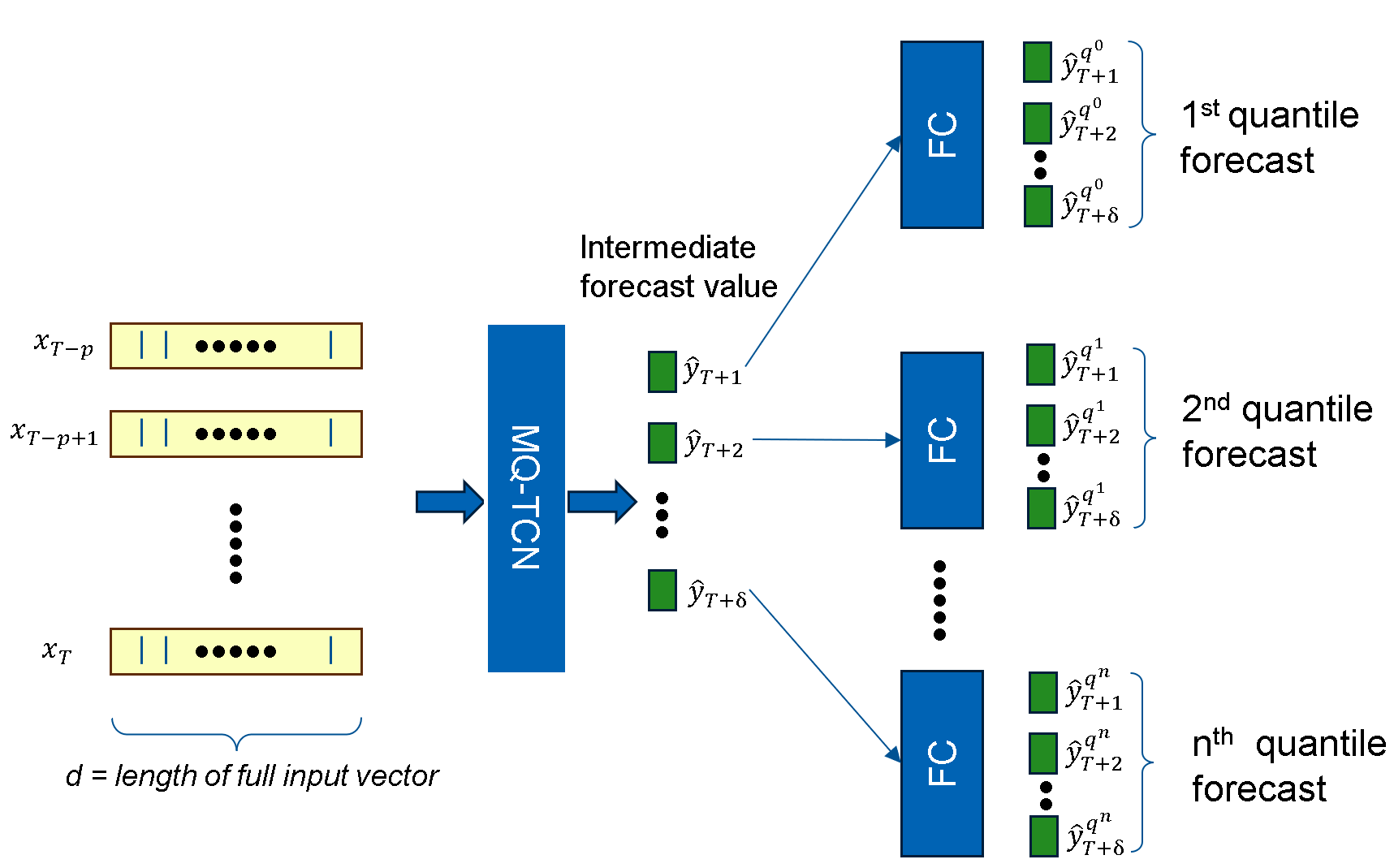}}
 \caption{Multi-Quantile Forecasting using TCN.}
 \label{fig:mq_tcn_architecture}
\end{figure}

Instead of single-point forecasting, we propose a TCN architecture tailored for quantile regression \cite{b27} in e-mobility demand forecasting in Figure \ref{fig:mq_tcn_architecture}. The TCN processes a sequence of inputs \( x_{T-p}, x_{T-p+1}, \ldots, x_T \), where \( p \) denotes the number of past observations and \( d \) represents the full length of the input vector. The network's output serves as an intermediate forecast value, which is then input into subsequent fully connected (FC) layers. Then, each FC layer in the model specializes in predicting a distinct quantile of the forecast distribution. Specifically, the output for the first quantile is given by \( \hat{y}_{T+1}^{q^0}, \hat{y}_{T+2}^{q^0}, \ldots, \hat{y}_{T+\delta}^{q^0} \); for the second quantile by \( \hat{y}_{T+1}^{q^1}, \hat{y}_{T+2}^{q^1}, \ldots, \hat{y}_{T+\delta}^{q^1} \); and so on, up to the \( n \)-th quantile.
The TCN model is trained using the Pinball Loss function \( L_{PB} \), which is designed to optimally penalize the forecast errors for each quantile level \( q \) as follows:
\begin{multline}
L_{PB}(q, y_{T+i}, \hat{y}_{T+i}^{q}) = \\
\begin{cases} 
(1-q) \cdot (y_{T+i} - \hat{y}_{T+i}^{q}) & \text{if } \hat{y}_{T+i}^{q} \geq y_{T+i}, \\
q \cdot (y_{T+i} - \hat{y}_{T+i}^{q}) & \text{if } \hat{y}_{T+i}^{q} < y_{T+i},
\end{cases}
\label{eq:pinball_loss}
\end{multline}
for each forecast horizon \( i=1,2,\ldots,\delta \) and corresponding quantile level \( q \). This function ensures that the quantile forecasts are not only centered but also capture the inherent variability of the future values, providing a comprehensive probabilistic forecast which is crucial for decision-making in dynamic and demand-responsive environments such as e-mobility services.
\subsection{Transfer Forecasting Knowledge with Inductive TL }
TL transfers knowledge from one EV load forecasting model trained on abundant data to forecast load on geographically separated charging sites with limited data. Based on past studies and scientific literature \cite{Zhuang2019ACS}, we define TL and its related terms as follows:

If ${x_{t}^{d}}$ is a sample data point at time $t$ with $d$ non-predictive features, a domain \emph{D} can be characterized by (i) feature space, $\chi^{d+1}$ (including non-predictive and target feature space) and (ii) marginal probability distribution \begin{math}P(X)\end{math} with $X= \left \{ x\mid x_{i}\in \chi,i=1......, d \right \}$. Domain \emph{D}  in TL can be denoted as  $D=\left \{ \chi,P(X) \right \}$ . A task, $\mathcal{T}$, can be defined via label space $\mathcal{Y}$ and objective function $ f\left (.\right )$ can be denoted as, $\mathcal{T}=\left \{ \mathcal{Y},f\left (.\right ) \right \}$  in where $f\left (.\right )$ can be only learned from training data $\left ( \left \{ x_{i},y_{i} \right \} \mid x_{i}\in X \quad and \quad y_{i}\in Y\right )$. A generic definition of inductive TL is as follows:

\textit{Definition (Inductive Transfer Learning): Given a source domain $D_{sd}$ and a learning task $\mathcal{T}_{sd}$, and a target domain $D_{td}$  and learning task $\mathcal{T}_{td}$, the goal of inductive TL aims to assist improve
the learning of the target predictive function  $f_{td}\left ( . \right )$  in $D_{td}$ using the knowledge in $D_{sd}$ and $\mathcal{T}_{sd}$, where
 $\mathcal{T}_{sd}$    $\neq $  $\mathcal{T}_{td}$}.
 
In accordance with the above formal definition, in our research experiment, our target load forecasting function is a non-linear deep learning model as described in the previous section under the name of MQ-TCN. The source and target domain feature spaces can be denoted as $\chi_{sd}\in \mathbb{R}^{(T\times d_{sd})}$ and $\chi_{td}\in \mathbb{R}^{(T\times d_{td})}$, respectively, where $(T\times d)$ is the optimal length of the input sequence values of individual domains, known as the lookback window used to train MQ-TCN model. The marginal probability distributions of the source and target variables (future energy consumption) are different: $P_{sd}(X)\neq P_{td}(Y)$. Both the source and target domain data in our research come from geographically separated EV charging sites and are not identical ($D_{sd}\neq D_{td}$) in terms of data distribution. The energy consumption profile of the EV user often varies from area to area, even within the same country, due to different lifestyles, economic status, and other factors. The source ($\mathcal{Y}_{sd}$) and target ($\mathcal{Y}_{td}$) label spaces in the case of energy forecasting model depend on the length of the forecast horizons. As the predictive function $ f\left (.\right )$ is responsible for transferring knowledge in TL settings, but $ f\left (.\right )$ is learned by providing training data from geographically separated EV charging sites. Therefore, source and target forecasting functions are also not equal: $ f_{sd}\left (.\right )\neq f_{td}\left (.\right ) $. In our research, we investigate both homogeneous TL ( $\chi_{sd}= \chi_{td}$ ; $\mathcal{T}_{sd}(Y)= \mathcal{T}_{td}(Y)$) and heterogeneous TL ( $\chi_{sd}\neq \chi_{td}$; $\mathcal{T}_{sd}(Y)\neq \mathcal{T}_{td}(Y)$) in where source and target features space can be both (same or different) and also for a constant source label space, forecasting horizons are variable (both same and different) for target task ($\mathcal{T}_{td}$) in various experimental settings.

We have applied a `head replacement' approach in our inductive TL experiments, as our TL based experiments are homogeneous and heterogeneous. In heterogeneous TL experiments, the label spaces of the source and target domains are different. However, this approach enables flexibility to make experiments with target domain (geographically separated charging infrastructure) load forecasting tasks of various time horizons ($\mathcal{T}_{td} \in \mathbb{R}^{1,4,24}$).

\section{Experimental Setup and Evaluation}
\label{sec:experimentalEvaluation}
This section give details about datasets, experimental setup and evaluation measure used in our research.

\subsection{Data}

\begin{figure*}[!htbp] 
  \centering
  \begin{subfigure}{0.58\textwidth} 
    \includegraphics[width=\linewidth]{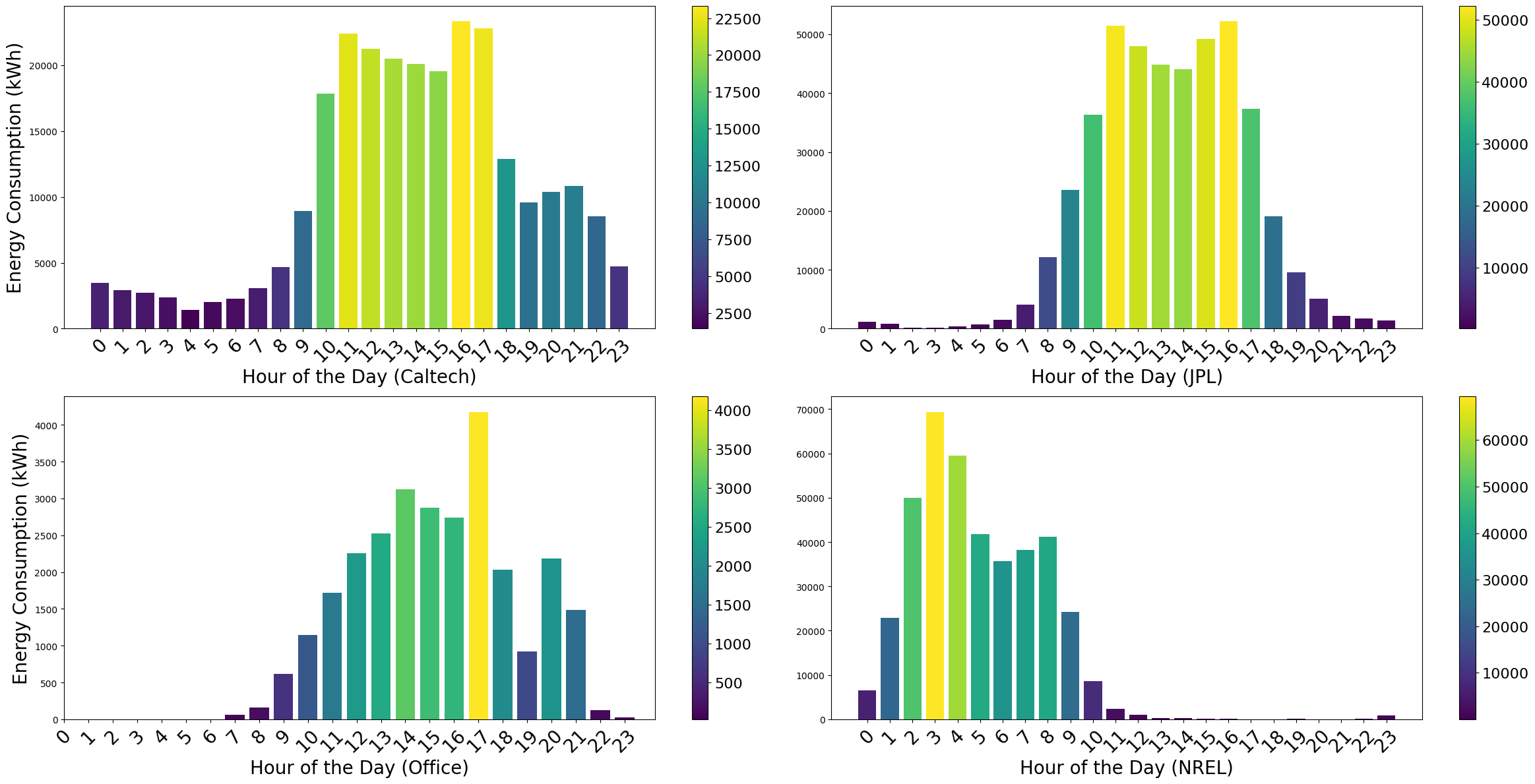}
    \caption{Location-wise Hourly Energy Consumption (KWh).}
    \label{fig:hourly_energy_consumption}
  \end{subfigure}
  \hfill
  \begin{subfigure}{0.39\textwidth} 
    \includegraphics[width=\linewidth]{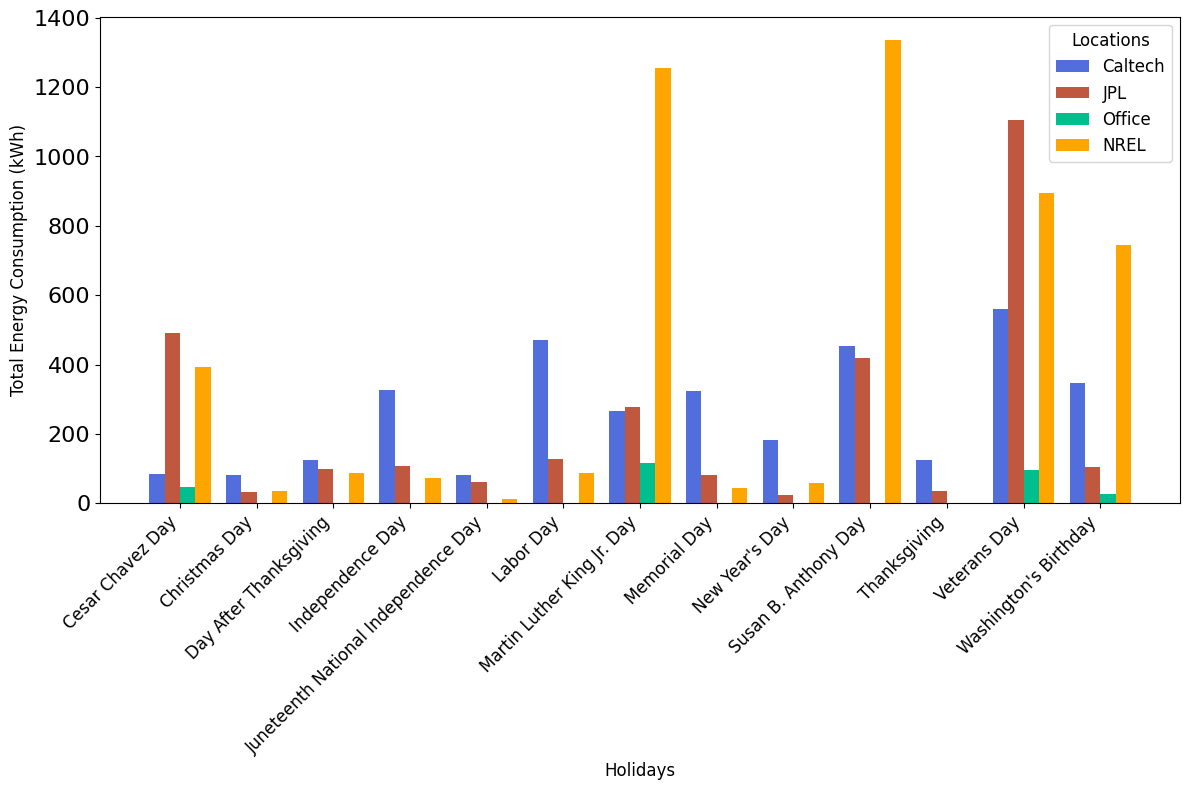}
    \caption{Location-wise Holiday's Energy Consumption (KWh).}
    \label{fig:holiday_energy_consumption}
  \end{subfigure}

  \vspace{2mm} 

  \begin{subfigure}{\textwidth} 
    \centering
    \includegraphics[width=0.75\linewidth, height = 2.2in]{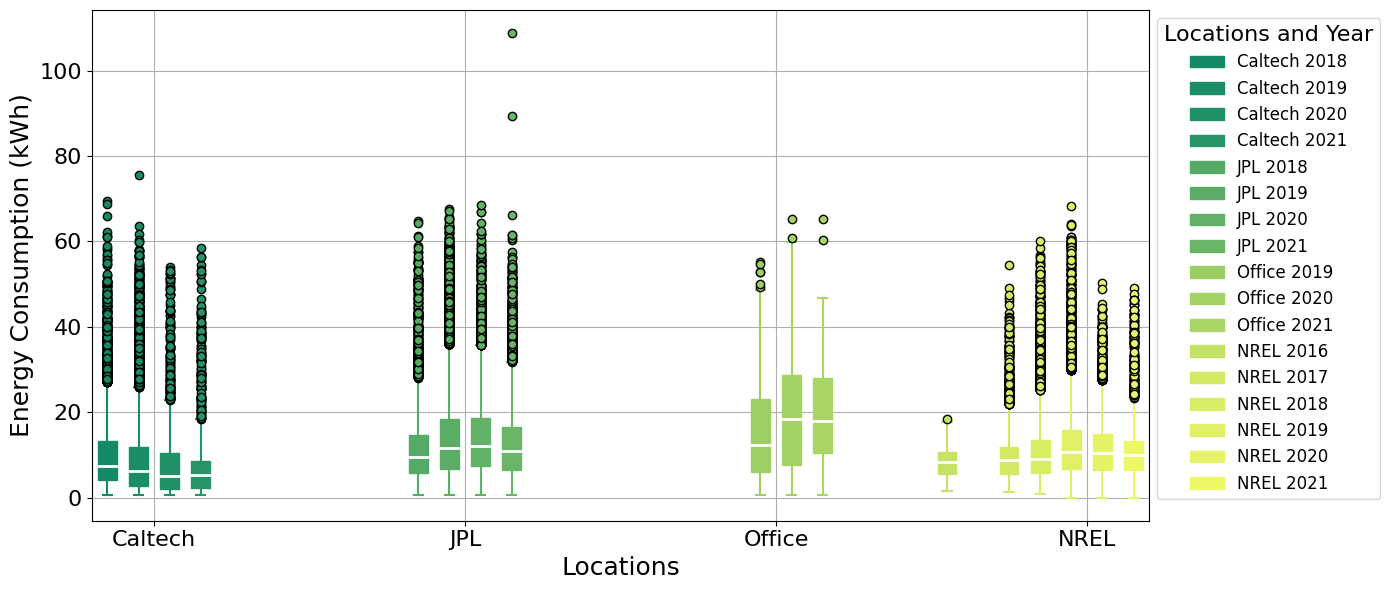}
    \caption{Location-wise Annual Energy Consumption (KWh) Boxplots.}
    \label{fig:yearwise_energy_boxplot}
  \end{subfigure}

  \caption{Energy Consumption Trends and Comparisons for All Locations} 
  \label{fig:energy_consumption_all_locations}
\end{figure*}

Table \ref{tab:datasetStat} provides detailed statistics about ACN and NREL datasets. ACN contains historical EV charging session information from the California Institute of Technology (Caltech), Jet Propulsion Laboratory (JPL), and a small office workplace (Office-1) \cite{lee_acndata_2019}. On the other hand, the NREL dataset represented historical charging session data of the National Renewable Energy Laboratory (NREL). The table shows Caltech has 54 charging stations with 13 different data features. JPL has slightly fewer stations than Caltech but more charging sessions. With only 8 stations, the Office-1 location has significantly fewer sessions than the other locations. Compared to the other three charging locations, NREL has the most stations, 141, and the highest number of sessions over the most extended period of approximately 59 months \cite{Neuman2021WorkplaceCD}. This site includes 16 data features, indicating a broader data collection and features than other sites.
\begin{table}[ht]
\centering
\caption{E-Mobility Dataset Statistics}
\setlength{\tabcolsep}{1.8pt} 
\renewcommand{\arraystretch}{1.2}
\footnotesize
\begin{tabular}{
>
{\centering\arraybackslash}p{1.1cm}|>{\centering\arraybackslash}p{2.2cm}|>{\centering\arraybackslash}p{2.2cm}|>{\centering\arraybackslash}p{1.2cm}|>{\centering\arraybackslash}p{1cm}}
\toprule
\textbf{\shortstack{Charging \\ Sites}} 
& \textbf{\shortstack{\# of\\ Charging Stations}} & \textbf{\shortstack{\# of\\ Charging Sessions}} & \textbf{\shortstack{Duration\\ (Months)}} 
& \textbf{\shortstack{\# of\\ Features}} \\
\midrule
Caltech & 54 & 31424 &  40 & 13 \\ 
JPL     & 50 & 33638 & 35 & 13 \\
Office-1  & 8  & 1683 & 30 & 13 \\ 
NREL  & 141  & 40979 & 59 & 16 \\
\bottomrule
\end{tabular}
\label{tab:datasetStat}
\end{table}
Figures \ref{fig:hourly_energy_consumption}, \ref{fig:holiday_energy_consumption}, and \ref{fig:yearwise_energy_boxplot} provide a comprehensive view of energy consumption trends at Caltech, JPL, Office-1, and NREL locations. In Figure \ref{fig:hourly_energy_consumption}, Caltech shows peak energy usage during daytime hours, JPL in the early afternoon, the Office-1 at midday, and NREL unusually peaks in the early morning. Figure \ref{fig:holiday_energy_consumption} illustrates varying holiday energy consumption with notable spikes at NREL during Susan B. Anthony Day and Martin Luther King Jr. Day, while JPL peaks on Veterans Day. On the other hand, the Office-1 consumes very little energy during holidays. Lastly, Figure \ref{fig:yearwise_energy_boxplot} displays annual energy usage, highlighting NREL's significant variability and the Office-1's consistent consumption across years, allowing for detailed analysis of temporal energy patterns across sites.\\
We prepared our datasets by addressing missing values where feasible to maintain consistent features across all locations. Anomalies were removed using Facebook Prophet \cite{Taylor2018ForecastingAS}, and domain knowledge. Hourly data samples were standardized for uniformity. We utilized sine-cosine transformations to capture the cyclic nature of temporal data. Numerical features were normalized using a Min-Max scaler method. Using an encoder-based dimensionality reduction technique, we translated a high-dimensional categorical feature such as `StationID' from 186 unique values to only 30 unique values of lower dimension.
 For low-dimensional categorical features, we applied different encoding strategies based on the number of unique categories: for features with 2 to 5 unique values, we used one-hot encoding. For features with 5 to 10 unique values, we used embedding layers in our DNNs to improve model performance. To avoid disruptions caused by the COVID-19 pandemic, we curated the data to include records only until March 2020.
\begin{figure*}[htbp]
  \centering
  \resizebox{0.9\linewidth}{!}{\includegraphics[]{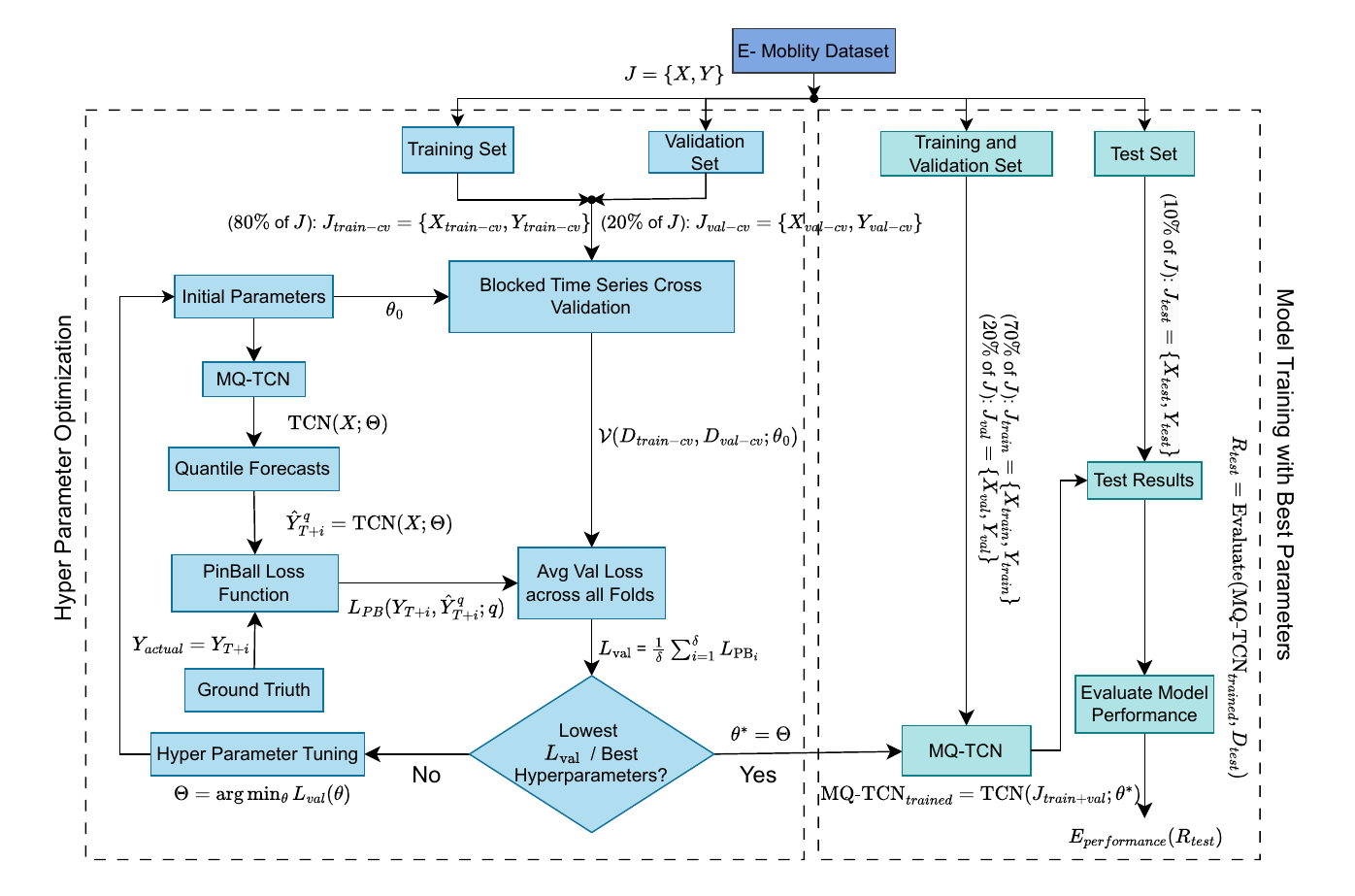}}
  \caption{Experimental Design for MQ-TCN Forecasting in E-Mobility Applications.}
  \label{fig:exp_design}
\end{figure*}

\begin{table*}[ht] 
    \centering 
    \caption{Forecasting Performance Comparison without Transfer Learning (Source Domain).} 
    \setlength{\tabcolsep}{2.8pt} 
    \renewcommand{\arraystretch}{0.8} 
    \footnotesize 
\begin{tabular}{
    >{\centering\arraybackslash}p{1.2cm}
    >{\centering\arraybackslash}p{0.65cm}
    >{\centering\arraybackslash}p{0.8cm}
    >{\centering\arraybackslash}p{1.0cm}
    >{\centering\arraybackslash}p{0.8cm}|   
    >{\centering\arraybackslash}p{1.3cm}
    >{\centering\arraybackslash}p{1.0cm}|
    >{\centering\arraybackslash}p{1.2cm}
    >{\centering\arraybackslash}p{0.65cm}
    >{\centering\arraybackslash}p{1.0cm}
    >{\centering\arraybackslash}p{0.8cm}
    >{\centering\arraybackslash}p{1cm}
} 

        \toprule
 & \multicolumn{3}{c}{\textbf{Caltech}} & \multicolumn{3}{c}{\textbf{}} & \multicolumn{5}{c}{\textbf{JPL}} \\ 
\cmidrule(r){1-5} \cmidrule(lr){6-7} \cmidrule(l){8-12}
\textbf{Model} & \textbf{PICP} & \textbf{PinBall Loss} & \textbf{WS} & \textbf{ND} & \textbf{Lookback Time} & \textbf{Forecast Horizon} & \textbf{Model} & \textbf{PICP} & \textbf{PinBall Loss} & \textbf{WS} & \textbf{ND} \\ 
\midrule
\multirow{3}{*}{MQ-TCN} & \multirow{3}{*}{80.19} & \multirow{3}{*}{3.38} & \multirow{3}{*}{38.90} & \multirow{3}{*}{0.3523} & \multirow{3}{*}{168} & \multirow{3}{*}{24} & MQ-TCN & \textbf{93.62} & 2.28 & 21.55 & 0.2226 \\
  &  &  &  & & & & XGBoost & 64.69 & \textbf{1.82} & 19.58 & \textbf{0.1770} \\
   &  &  & & & & & DeepAR & n/a & n/a & n/a & \textbf{0.1770} \\
\midrule

  MQ-TCN & \textbf{84.93} & 3.47 & 37.04 & 0.3602 & \multirow{3}{*}{168} & \multirow{3}{*}{48} & \multirow{3}{*}{MQ-TCN} & \multirow{3}{*}{75.50} & \multirow{3}{*}{2.35} & \multirow{3}{*}{25.04} & \multirow{3}{*}{0.2266} \\
  XGBoost & 78.96 & 3.13 & 33.41 & 0.3254 & & & & & & \\
  DeepAR & n/a & n/a & n/a & \textbf{0.2955} & & & & & & \\
\midrule
  MQ-TCN & 82.79 & 3.39 & 38.94 & 0.3649 & 336 & 24 & MQ-TCN & 87.85 & 2.64 & 20.93 & 0.2573 \\
\midrule
  MQ-TCN & 80.90 & \textbf{3.04} & \textbf{32.37} & 0.3261 & 336 & 48 & MQ-TCN & 88.32 & 1.89 & \textbf{18.16} & 0.1814 \\
\bottomrule
    \end{tabular}
\label{tab:source_model_table}
\end{table*}
\subsection{Evaluation Metrics}

In our experiments, we evaluated model performance using several evaluation metrics: PICP, Pinball Loss, WS Score, and Normalized Deviation (ND). Pinball Loss is described in Equation \ref{eq:pinball_loss}. PICP, WS, and ND are defined as:
\begin{equation}
\text{PICP} = \frac{1}{N} \sum_{i=1}^{N} \varepsilon_i
\end{equation}
where the variable \( \varepsilon_i \) is defined as follows:
\begin{equation}
\varepsilon_i =
\begin{cases}
1, & \text{if } y_{T+i} \in [L_{T+i, q}, U_{T+i, q}], \\
0, & \text{if } y_{T+i} \notin [L_{T+i, q}, U_{T+i, q}].
\end{cases}
\end{equation}

\begin{equation}
\begin{aligned}
WS = \sum_{i=1}^{N} \begin{cases}
\gamma, & L_{T+i,q} \leq y_{T+i} \leq U_{T+i,q} \\
\gamma + \frac{2 \left (L_{T+i,q} - y_{T+i} \right)}{\alpha}, & y_{T+i} < L_{T+i,q} \\
\gamma + \frac{2 \left (y_{T+i} - U_{T+i,q} \right)}{\alpha}, & y_{T+i} > U_{T+i,q}
\end{cases}
\end{aligned}
\end{equation}

\begin{equation}
\text{ND} = \frac{\sum_{i=1}^{N} | y_{T+i} - \hat{y}_{T+i}|}{\sum_{i=1}^{N} |y_{T+i}|}
\end{equation}
Here, the variable \( \varepsilon_i \) is a Boolean variable that indicates the coverage behavior of prediction intervals (PIs). \(y_{T+i}\) denotes the actual value at time \(T+i\), \(\hat{y}_{T+i}\) denotes the forecasted value at the same timestamp, and \(N\) is the number of periods over which the forecast is evaluated. \( \gamma \) is the interval width and \(\alpha\) = 0.15 in our settings.

\subsection{Multivariate Multi-Step Probabilistic Load Forecasting}

In this section, we described results and analysis in response to research question 1 by evaluating the proposed MQ-TCN model on only Caltech and JPL datasets.

In addition to our proposed MQ-TCN model, we also explored two well-known forecasting models, namely XGboost and DeepAR, as baselines. Past research works did not consider high cardinal categorical variables in conjunction with temporal and numerical features, particularly concerning multivariate, multi-step load forecasting of charging station using deep learning models \cite{b1} \cite{b2}. As our dataset is heterogeneous time-series nature, we have additional engineering challenges in adapting traditional XGBoost and DeepAR architectures. We modified the traditional XGBoost architecture, which is suitable only for point-based forecasting, to quantile regression based on probabilistic forecasting to compare with MQ-TCN. In addition, we needed to modify stride size in standard DeepAR architecture to fit our experimental heterogeneous time-series dataset. We divided our entire experimental datasets into training, validation, and test sets, as shown in Figure \ref{fig:exp_design}. To evaluate the model's performance, we start by creating a test set. This set consists of the last 10\% \( J_{test}\) of the data \( J(X, Y)\) and is set aside to evaluate the model. The remaining data is then divided into several folds \( f_{fold}\) for time-series blocked cross-validation. Each fold contains 80\% \( J_{train-cv}\) of the data for training and 20\% \( J_{val-cv}\) for validation. We calculate the average validation loss over all folds \( f_{fold}\) and use the loss function \( L_{PB}\) to identify the best hyperparameter set. With the best hyperparameters, we retrain the model using 90\% \( J_{train+val}\) of \( J(X, Y)\) without shuffling. Finally, the MQ-TCN model is trained and evaluated using the previously separated \( J_{test}\) test set.

Eight load forecasting models were developed and trained on Caltech and JPL datasets with varying settings for three quantile levels (0.05, 0.50, 0.90). These models were based on MQ-TCN, XGBoost, and DeepAR, and utilized various lookback periods (336 hours, 168 hours) and forecasting horizons (48 hours, 24 hours). Since Office-1 has a very limited number of samples, we excluded it here. For training XGBoost and DeepAR, we select and keep settings similar to the best performing MQ-TCN model based on the highest PICP score for comparison. A Bayesian optimization based method named Tree Prazen Sampler \cite{b33} was used to search for hyperparameters for all models and settings. We compared our proposed MQ-TCN and XGBoost models using PICP, Pinball Loss, and WS and compared MQ-TCN and DeepAR using ND. The optimization was based on ND in the case of DeepAR, and for our case, we considered 50\% quantile predictions to make the comparison.\\
\textbf{Results and Detailed Analysis:} Experimental results displayed in Table \ref{tab:source_model_table} indicate a significant improvement in the performance of our MQ-TCN model, compared to the XGBoost baseline, across various lookback periods and forecasting horizons. Our MQ-TCN model achieved higher coverage rates, particularly for site Caltech, where the optimal model utilized 168 hours of historical data and forecasted for 48 hours. The XGBoost baseline had a coverage rate of 78.96\%, while the MQ-TCN model achieved a PICP of 84.93\%, which is an impressive 5.97\% coverage enhancement. We observe improvements of 28.93\% multi-step forecasting (24 hours) at site JPL with our MQ-TCN over XGBoost. However, DeepAR outperform MQ-TCN and XGBoost in terms of ND at the Caltech location. A sample load forecasting of 168 hours for the source sites Caltech and JPL are shown in Figure \ref{fig:source_model_caltech} and \ref{fig:source_model_JPL}.
\begin{table}[t!] 
    \centering 
    \caption{Results of Load Forecasting at Office-1 Site with Inductive TL (Target Domain). } 
    \setlength{\tabcolsep}{0.1pt} 
    \renewcommand{\arraystretch}{0.6} 
    \footnotesize 
\begin{tabular}{  
    >{\centering\arraybackslash}p{1.2cm}|
    >{\centering\arraybackslash}p{1.1cm}|
    >{\centering\arraybackslash}p{1.2cm}
    >{\centering\arraybackslash}p{1.2cm}|
    >{\centering\arraybackslash}p{0.9cm}
    >{\centering\arraybackslash}p{0.9cm}
    >{\centering\arraybackslash}p{0.8cm}
    >{\centering\arraybackslash}p{1cm}
} 

        \toprule
 \multicolumn{8}{c}{\textbf{Office-1}} \\ 
\midrule
\textbf{Model} & \textbf{Data Size} & \textbf{Lookback Time} & \textbf{Forecast Horizon} & \textbf{PICP} & \textbf{PinBall Loss} & \textbf{WS} & \textbf{ND} \\ 
\midrule
MQ-TCN (JPL-Source) & Full Data & 168 & 24 & 93.62 & 2.28 & 21.55 & 0.2226 \\
\midrule
XGBoost & & \multirow{2}{*}{168} & \multirow{2}{*}{24} & 53.03 & 0.35 & 4.27 & 0.4430\\
DeepAR & & & & n/a & n/a & n/a & 1.0620 \\
\cmidrule(r){1-1}\cmidrule(lr){3-4}\cmidrule(l){5-8}
XGBoost & & \multirow{2}{*}{72} & \multirow{2}{*}{24} & \text{60.45} & 0.35 & 3.82 & 0.4305 \\
DeepAR & Full Data & & & n/a & n/a & n/a & 1.0587 \\
\cmidrule(r){1-1}\cmidrule(lr){3-4}\cmidrule(l){5-8}
XGBoost & & 24 & 4 & 41.02 & \text{0.33} & \text{3.48} & \text{0.4291} \\
DeepAR & & 24 & 1 & n/a & n/a & n/a & 1.0810 \\
\cmidrule(r){1-1}\cmidrule(lr){3-4}\cmidrule(l){5-8}
XGBoost & & 24 & 1 & 37.24 & \text{0.33} & \text{3.62} & \text{0.4236} \\
  \midrule
\multirow{2}{*}{MQ-TCN} & \multirow{2}{*}{2 weeks} & 24 & 4 & \text{87.50} & \text{0.89} & \text{8.14} & \text{0.4641} \\
  & & 24 & 1 & 84.38 & \text{0.90} & \text{8.83} & \text{0.4725} \\
  \midrule
 \multirow{3}{*}{MQ-TCN} & & 72 & 24 & 87.30 & \text{0.66} & 6.18 & \text{0.4398} \\
 & 1 month & 24 & 4 & \textbf{91.04} & 0.74 & \text{6.03} & 0.6716 \\
 & & 24 & 1 & \textbf{86.76} & \text{0.73} & \text{6.21} & \text{0.5372} \\
  \midrule
 & & 168 & 24 & 80.71 & \textbf{0.40} & \textbf{3.77} & \textbf{0.4108} \\
  \multirow{2}{*}{MQ-TCN} & \multirow{2}{*}{3 months} & 72 & 24 & 86.12 & 0.61 & 5.07 & 0.5862  \\
 & & 24 & 4 & 81.02 & 0.49 & 5.14 & 0.4695 \\
 & & 24 & 1 & 77.88 & \text{0.49} & \text{5.44} & \text{0.4509} \\
   \midrule
& & 168 & 24 &  \textbf{89.23} & 0.45 & 4.67 & 0.4489 \\
\multirow{2}{*}{MQ-TCN} & \multirow{2}{*}{6 months} & 72 & 24 & 88.55 & \text{0.42} & \text{4.25} & \text{0.4262} \\
 & & 24 & 4 & 84.14 & 0.45 & 4.37 & 0.4434 \\
 & & 24 & 1 & 59.08 & \textbf{0.44} & \textbf{4.44} & \textbf{0.4353} \\
\bottomrule
\end{tabular} 
\label{tab:officeForecast}
\end{table}

\begin{figure}[htbp]
  \centering
  \resizebox{\linewidth}{!}{\includegraphics[width=1.4\textwidth]
  {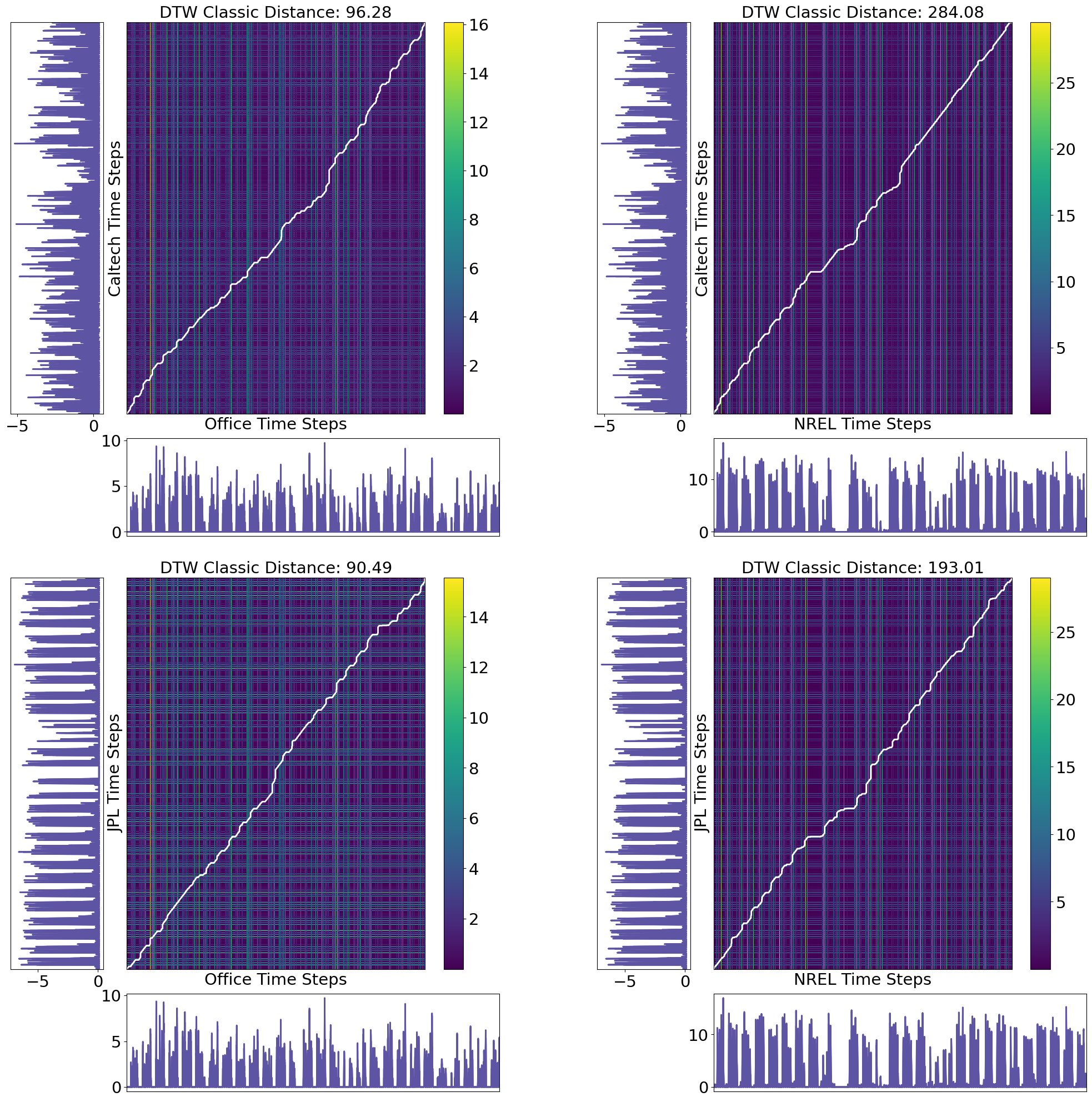}}
    \caption{Similarity Distance Measurement between Two EV Charging Sites Datasets using DTW method.}
  \label{fig:dtw_plot}
\end{figure}

\begin{table}[t!] 
    \centering 
    \caption{Results of Load Forecasting at NREL Site with Inductive TL (Target Domain).} 
    \setlength{\tabcolsep}{0.1pt} 
    \renewcommand{\arraystretch}{0.6} 
    \footnotesize 
\begin{tabular}{  
    >{\centering\arraybackslash}p{1.2cm}|
    >{\centering\arraybackslash}p{1.1cm}|
    >{\centering\arraybackslash}p{1.2cm}
    >{\centering\arraybackslash}p{1.2cm}|
    >{\centering\arraybackslash}p{0.9cm}
    >{\centering\arraybackslash}p{0.9cm}
    >{\centering\arraybackslash}p{0.8cm}
    >{\centering\arraybackslash}p{1cm}
} 

        \toprule
 \multicolumn{8}{c}{\textbf{NREL}} \\ 
\midrule
\textbf{Model} & \textbf{Data Size} & \textbf{Lookback Time} & \textbf{Forecast Horizon} & \textbf{PICP} & \textbf{PinBall Loss} & \textbf{WS} & \textbf{ND} \\ 
\midrule

MQ-TCN (JPL-Source) & Full Data & 168 & 24 & 93.62 & 2.28 & 21.55 & 0.2226 \\
\midrule
XGBoost & & \multirow{2}{*}{168} & \multirow{2}{*}{24} & 46.04 & 2.04 & 22.91 & 0.3112\\
DeepAR & & & & n/a & n/a & n/a & 1.0939 \\
\cmidrule(r){1-1}\cmidrule(lr){3-4}\cmidrule(l){5-8}
XGBoost & & \multirow{2}{*}{72} & \multirow{2}{*}{24} & 80.21& 1.82 & 18.74 & 0.2944 \\
DeepAR & Full Data & & & n/a & n/a & n/a & 1.0562 \\
\cmidrule(r){1-1}\cmidrule(lr){3-4}\cmidrule(l){5-8}
XGBoost & & 24 & 4 & 71.49 & \text{1.73} & \text{17.98} & \text{0.2681} \\
DeepAR & & 24 & 1 & n/a & n/a & n/a & 1.0520 \\
\cmidrule(r){1-1}\cmidrule(lr){3-4}\cmidrule(l){5-8}
XGBoost & & 24 & 1 & 78.65& \text{1.74} & \text{17.49} & \text{0.2718} \\
  \midrule
\multirow{2}{*}{MQ-TCN} & \multirow{2}{*}{2 weeks} & 24 & 4 & 77.42& \text{3.16} & \text{23.26} & \text{2.0426} \\
 & & 24 & 1 & \textbf{96.88} & \text{1.86} & \text{9.78} & \text{1.2854} \\
  \midrule
\multirow{3}{*}{MQ-TCN} & & 72 & 24 & \textbf{95.08}& 4.28 & 28.07 & 0.7647 \\
& 1 month & 24 & 4 & 82.09 & \textbf{1.92} & \textbf{20.07} & \textbf{0.5969} \\
& & 24 & 1 & 85.29& \textbf{1.01} & \textbf{9.31} & \textbf{0.3166} \\
\bottomrule
    \end{tabular} 
    \label{tab:nrelForecast}
\end{table}

\begin{figure*}[!htbp]
  \centering
  \begin{subfigure}[b]{0.49\textwidth}
    \centering
    \includegraphics[width=\linewidth]{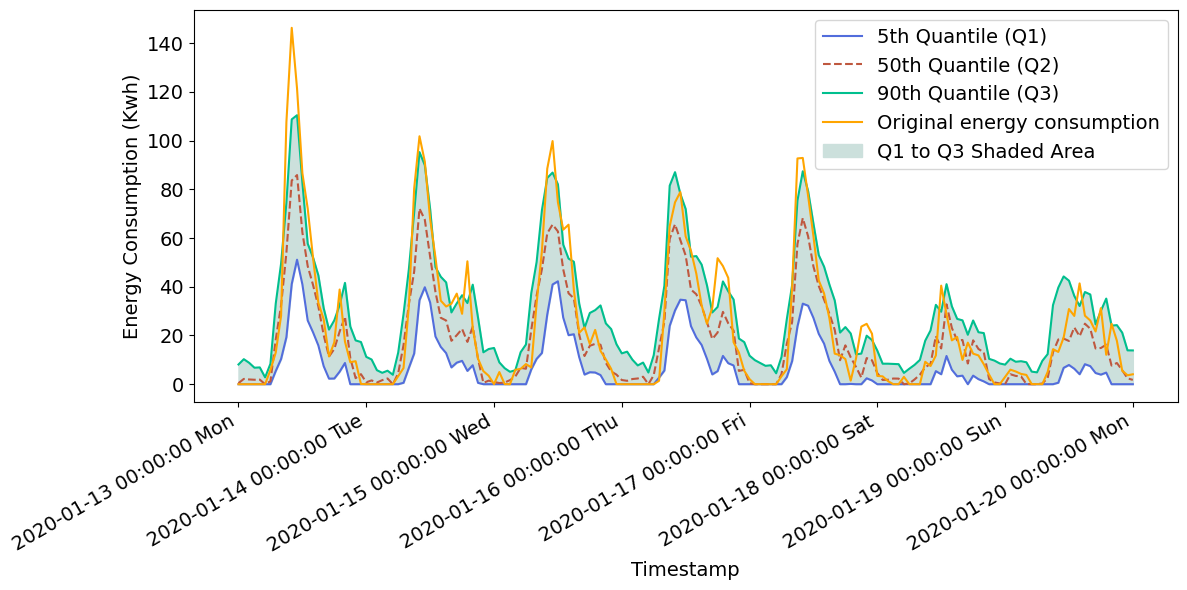}
    \caption{Sample Energy Consumption in a Week (168 Hours) for the \\ location Caltech.}
    \label{fig:source_model_caltech}
  \end{subfigure}%
  \begin{subfigure}[b]{0.49\textwidth}
    \centering
    \includegraphics[width=\linewidth]{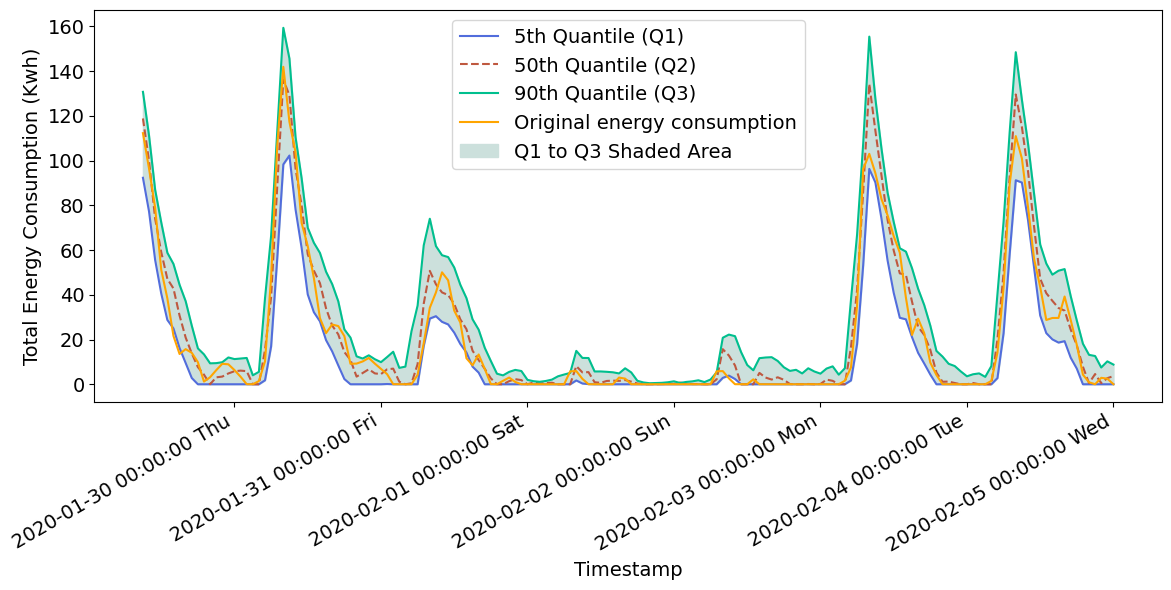}
    \caption{Sample Energy Consumption in a Week (168 Hours) for the \\ location JPL.}
    \label{fig:source_model_JPL}
  \end{subfigure}
  
  \begin{subfigure}[b]{0.49\textwidth}
    \centering
    \includegraphics[width=\linewidth]{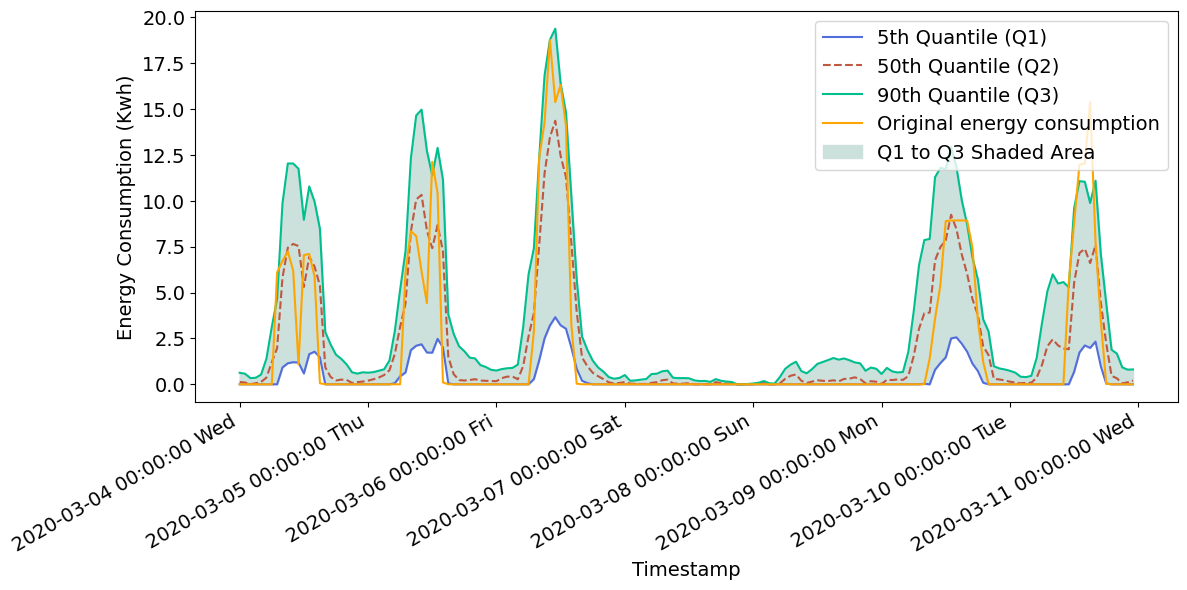}
    \caption{Sample Energy Consumption in a Week (168 Hours) for the \\ location Office-1 using TL from the location JPL.}
    \label{fig:target_model_Office}
  \end{subfigure}%
  \begin{subfigure}[b]{0.49\textwidth}
    \centering
    \includegraphics[width=\linewidth]{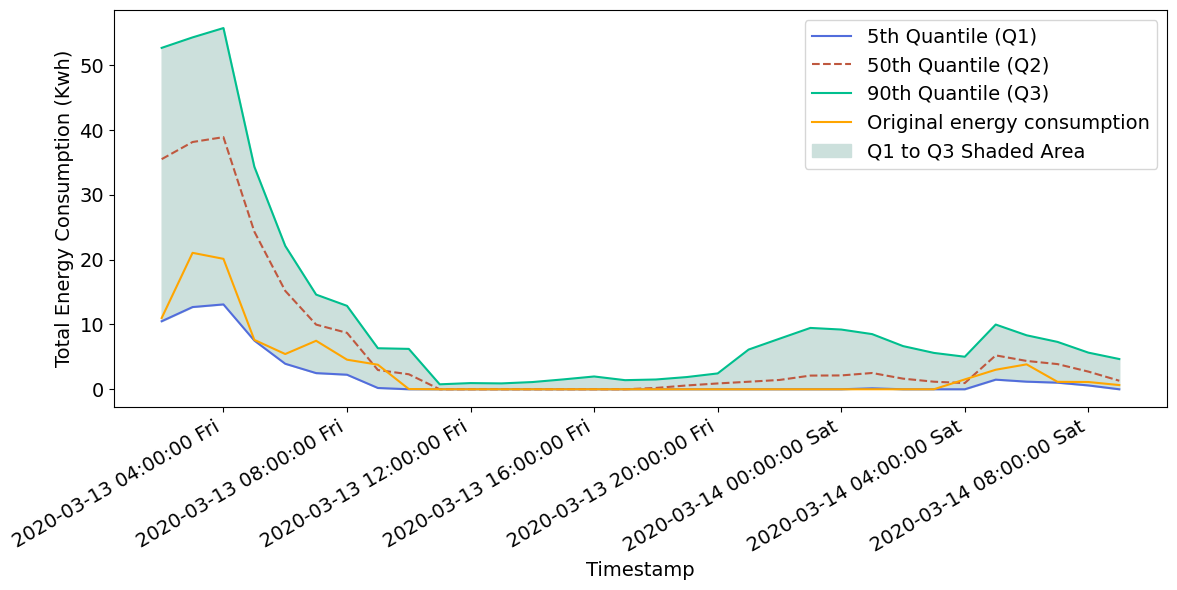}  
    \caption{Sample Energy Consumption in a Single Day (24 hours) for the \\ location Office-1 using TL from the location JPL.}
    \label{fig:target_model_NREL}
  \end{subfigure}
  \caption{Sample Forecasting of Source Domain (Caltech, JPL) and Target Domain (Office-1, NREL).}
\end{figure*}


\subsection{Location based Load Forecasting under Data Scarcity}
In response to research question 2, we conducted experiments in two stages, taking into account data from four EV charging stations: Caltech, JPL, Office-1, and NREL. Among these four locations, Caltech and JPL sites have more data samples than Office-1 and NREL, as shown in the table. Office-1 and NREL need more data samples to build efficient load models. It would be highly advantageous to reuse or transfer knowledge from pre-trained models (trained on initially large amounts of data, such as Caltech and JPL) to low-resource (limited data) scenarios like at Office-1 and NREL sites to forecast load.

In the first stage, we analyzed EV users' demographics and charging infrastructure type (public, private, or workplace). In addition, we quantify the `similarity' between any two charging sites out of four utilizing charging-related data with the `Dynamic-Time-Warping (DTW)' method illustrated in Figure \ref{fig:dtw_plot}. The choice to prioritize TL from datasets with lower DTW scores, \textit{`JPL-Office-1'} and \textit{`JPL-NREL'}, reflects their closer resemblance. Conversely, higher DTW scores were observed in \textit{`Caltech-Office-1' }and \textit{`Caltech-NREL' }pairs, indicating significant dissimilarities. Emphasizing datasets with inferior DTW scores enhances knowledge transfer efficacy, fostering improved model generalization and predictive accuracy \cite{b28}.

\begin{figure}[!htbp] 
  \centering
  \begin{subfigure}[b]{0.24\textwidth} 
    \includegraphics[width=\linewidth]{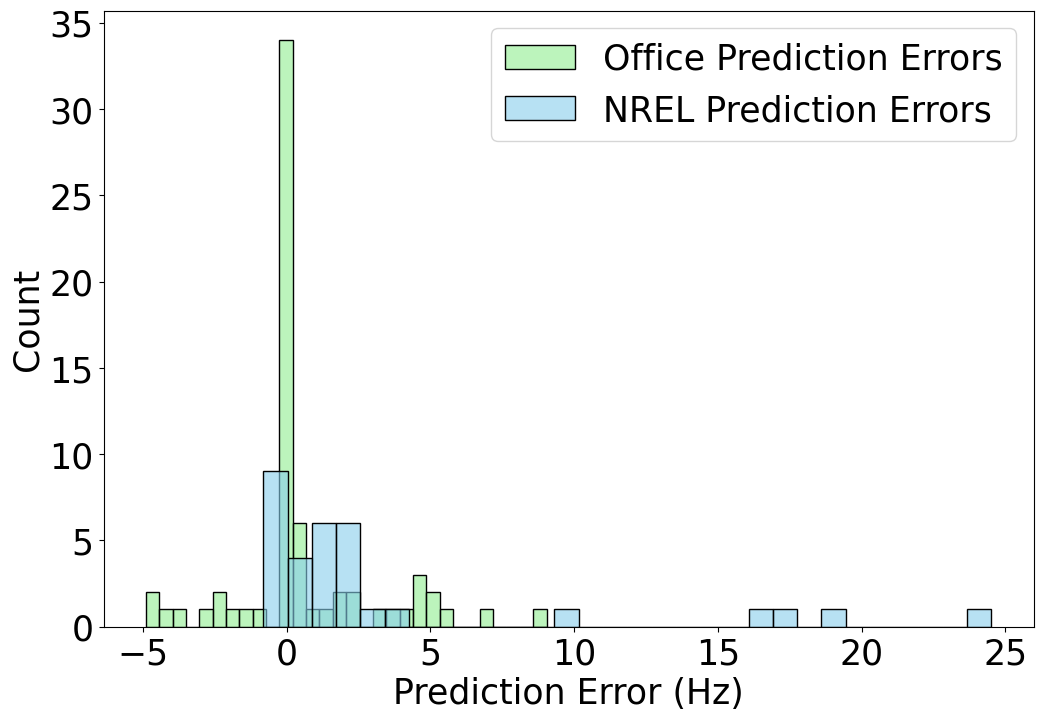}
    \caption{Comparison of Prediction Error Histograms.}
    \label{fig:histogram}
  \end{subfigure}
  \hfill
  \begin{subfigure}[b]{0.24\textwidth} 
    \includegraphics[width=\linewidth]{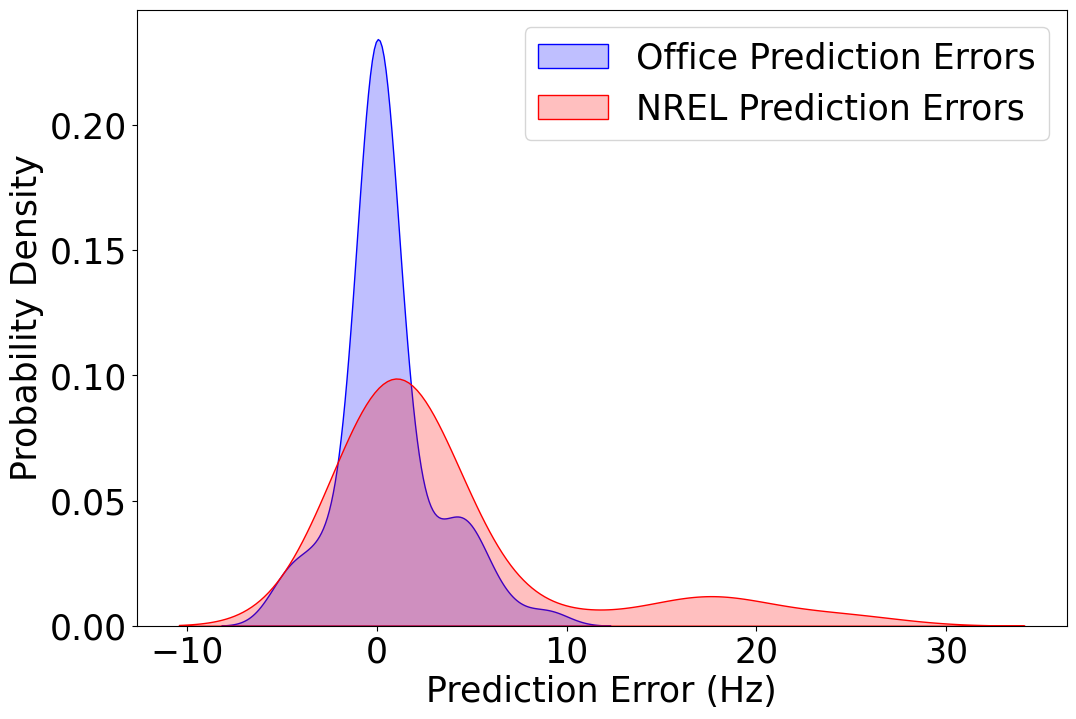}
    \caption{Comparison of Prediction Error PDFs.}
    \label{fig:kdeplot}
  \end{subfigure}
  \caption{Visual analysis of prediction errors between Office-1 and NREL sites (a) Histogram of prediction errors; (b) Kernel density estimate of prediction errors.}
  \label{fig:prediction_errors}
\end{figure}
A sample load forecasting of 1 week (168 hours) and 1 day (24 hours) for the target site Office-1 and NREL is shown in Figure \ref{fig:target_model_Office} and \ref{fig:target_model_NREL}. Also, Considering the optimal models for Office-1 and NREL based on the PICP, Figure \ref{fig:histogram} displays a histogram that quantifies the frequency of specific prediction error values. Unlike NREL's, it shows that Office-1's prediction errors are tightly clustered around zero. Figure \ref{fig:kdeplot} provides kernel density estimates of these prediction errors, highlighting their central tendencies and variances. It appears that the accuracy of Office-1's errors is higher and less variable as compared to NREL's errors, which are more dispersed.
  
In the second stage, we focus on optimizing knowledge transfer between similar charging sites, as discussed earlier. We utilized the best pre-trained model for JPL's location and aimed to transfer knowledge within the exact usage pattern for locations: Office-1 and NREL. We used the pre-trained MQ-TCN model as a feature extractor, removing the last layer and adding TCN blocks on top of it. Additionally, we added fully connected layers to predict each quantile. \\
\textbf{Results and Detailed Analysis:} Tables \ref{tab:officeForecast} and \ref{tab:nrelForecast} present a detailed comparison of forecasting performance at Office-1 and NREL using inductive TL across various models, lookback periods, and forecast horizons. We compare our result with baseline DeepAR and XGBoost models trained with complete data. Notably, with minimal data, our proposed MQ-TCN models exhibit robust performance across various data sizes, lookbacks, and forecasting horizons.

For the Office-1 site, XGBoost model, training with the entire dataset, we achieve a maximum PICP score of a maximum of 53.03\% during multi-step forecasting and 37.24\% across single-step forecasting scenarios. The performance of XGBoost generally shows a trend where shorter historical data periods and forecasting horizons correlate with poorer outcomes in terms of PICP and other metrics. XGBoost misses the very low forecasting values. DeepAR, trained with full data size for a one-day and four-hour forecast horizon, does not support single-step forecasting. Our proposed MQ-TCN outperforms DeepAR in terms of ND in all scenarios.

The MQ-TCN model, especially with a one-month dataset for Office-1 site, exhibits notable enhancements in PICP. It can reach as high as 91.04\% when using one day of historical data for forecasting four hours ahead in a multi-step forecast. In contrast, it achieves a PICP score of 86.76\% for single-step forecasting using the same dataset size. The mean PICP and ND scores of our proposed MQ-TCN are 83.36 and 0.4813, respectively. These scores are significantly better than the XGBoost's average PICP score of 47.935 and the DeepAR's average ND score of 1.067. Also, for NREL, we investigated how to transfer knowledge with a very limited amount of data, and we found that with a very limited amount of data, we achieved a PICP score of 96.88\% for single-step and 95.08\% for multi-step forecasting, which also outperform the baseline models.
\subsection{Uncertainty Handling and Impact of Variable Data Size, lookback Period and Forecast Horizon in TL }
\label{impact}
\begin{figure}[htbp]
  \centering
  \resizebox{\linewidth}{!}{\includegraphics[width=0.5\textwidth]{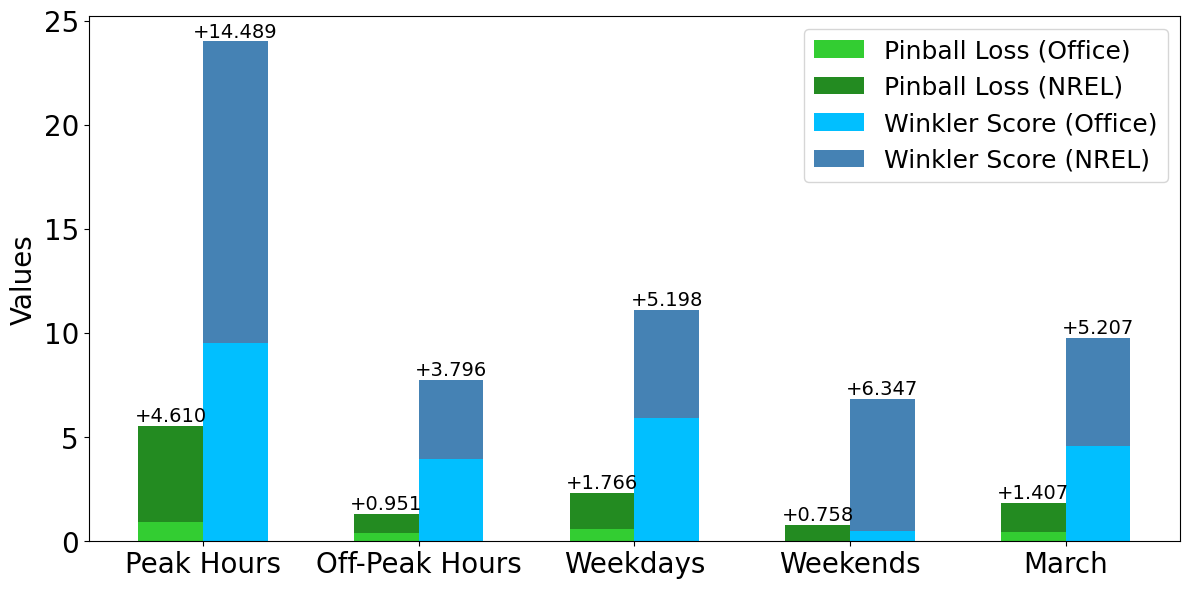}}
  \caption{Comparison of Predictive Precision and Interval Quality: Pinball Loss and Winkler Score for Office-1 and NREL}
  \label{fig:office_nrel_ev_met_bar}
\end{figure}

\begin{figure}[htbp]
  \centering
  \includegraphics[height=0.3\textheight]{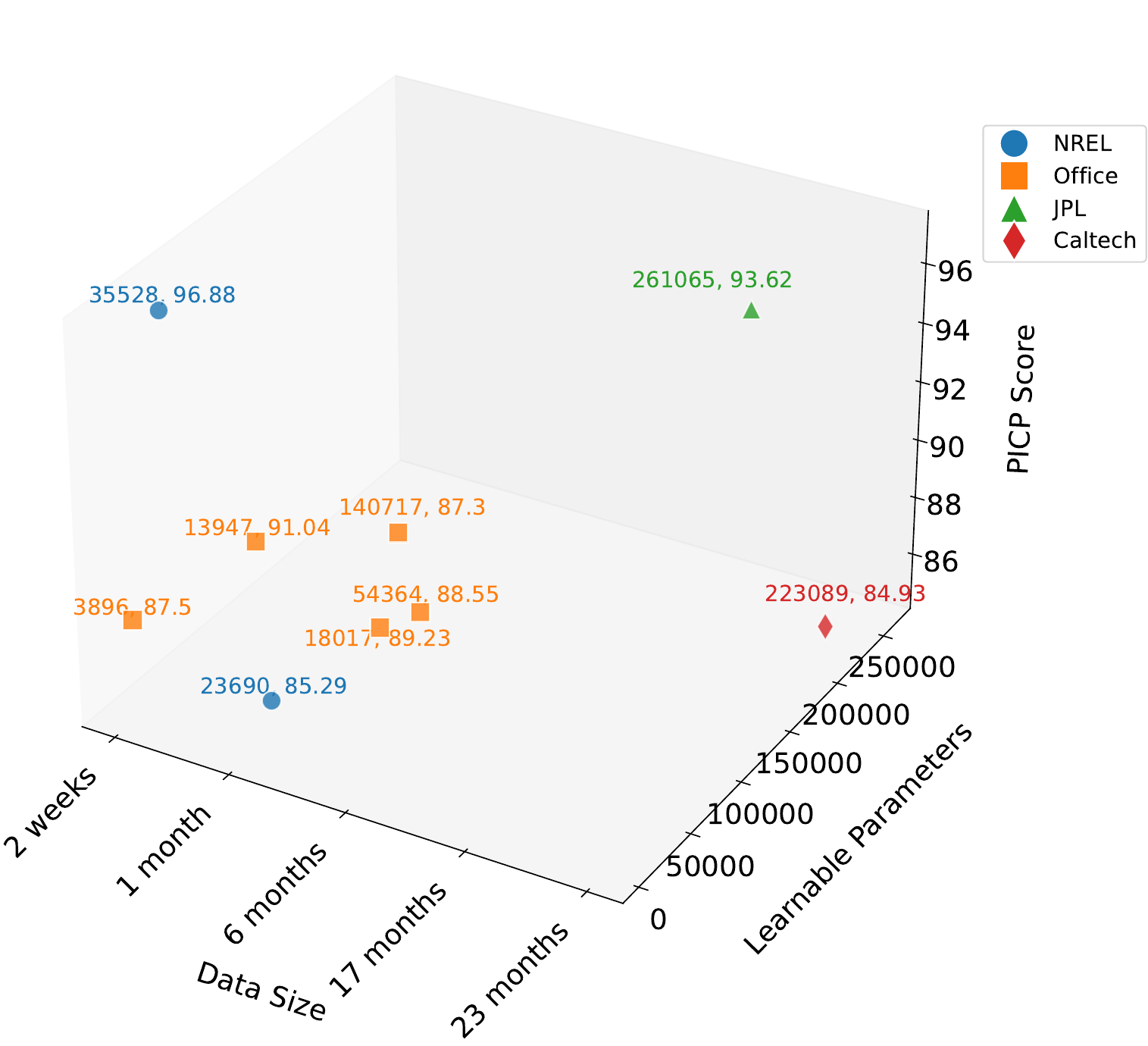}
  \caption{Comparison of PICP Scores and Learnable Parameters across Different Data Sizes and Locations}
  \label{fig:low_cost_3d_plot}
\end{figure}

In this section, we executed experiments using various data sizes to assess the impact of data size, lookback period, and forecast horizon on the performance of the MQ-TCN model and to estimate uncertainty. We started with different data sizes, such as 6 months, 3 months, 1 month and 2 weeks for Office-1, and 1 month and 2 weeks for NREL location. We tested the lookback (168, 72, 24) and future steps (24, 4, 1) accordingly for multi-step and single-step prediction.\\
\textbf{{Results and Detailed Analysis:}} The bar chart in Figure \ref{fig:office_nrel_ev_met_bar} presents information on the accuracy of energy consumption predictions in two different locations, namely Office-1 and NREL. Table \ref{tab:officeForecast} shows that Office-1 has higher reliability in its energy consumption predictions, as reflected by a robust PICP score of 89.23\%. This suggests that Office-1's forecasts are generally more accurate, making its energy management strategy potentially more reliable and actionable.
On the other hand, NREL achieved an impressive PICP score of 96.88\%, which might suggest superior forecasting but reveals a twist. NREL's larger WS indicates that its PI is wider, suggesting that while NREL's forecasts often capture the actual values, the range of these predictions is broad, pointing to less certainty about where the true values will fall within these ranges. The forecasting model for Office-1 is more precise, providing accurate predictions that make its energy management strategy more reliable. \\
We found no conclusive evidence of a relationship between lookback size, forecast horizon, and forecast accuracy. It was found that forecasting performance could be reduced or increased by increasing or decreasing the lookback period while keeping the forecast horizon constant. Similarly, with a fixed lookback period, increasing or decreasing the forecast horizon can increase or decrease forecast accuracy. As shown in Table \ref{tab:officeForecast}, at the Office-1 site, an increasing lookback period for a fixed day-ahead forecast horizon from 72 hours to 168 hours decreased forecast accuracy. \\
The scatter plot in Figure \ref{fig:low_cost_3d_plot} compares locations (NREL, Office-1, JPL, and Caltech) by data size, learnable parameters, and PICP score. NREL has PICP scores of 96.88 and 85.29, while the Office-1 ranges from 87.3 to 91.04. JPL scores 93.62, and Caltech scores 84.93. Notably, the target sites (Office-1 and NREL) have lower learnable parameters than the source sites (Caltech and JPL), with significantly higher learnable parameters. The plot demonstrates that even with very few learnable parameters in transfer learning (TL) settings, we can achieve comparable PICP scores. This reduction of the number of learnable parameters without comprising model's efficiency to make efficient forecasts significantly reduce high computation cost (computation time, cost for data collection, and maintenance etc.) \cite{thompson2020computational}.



\section{Conclusion and Future Work}
\label{sec:conclusion}
In our research, we successfully implemented and showed the applicability of our proposed MQ-TCN architecture for EV load forecasting tasks for multiple geographically separate EV charging sites under data scarcity. Based on the evaluation of the MQ-TCN model to transfer knowledge for load forecasting tasks among charging stations in TL settings, we described its effectiveness and adaptability. The MQ-TCN model provides load forecasting with minimal training data, precisely two weeks or 336 data instances, with high accuracy, reducing training time and, thus, computation cost. Though MQ-TCN achieved impressive results in transferring knowledge and showed adaptability to learn under data scarcity, it failed in some novel situations. We plan to explore this problem in the future, considering continual representation learning and the addition of meta-information.
\section*{Acknowledgment}
This work results from the project SALM
(01IS20098B) funded by BMBF (German Federal Ministry of Education and Research).
\addcontentsline{toc}{chapter}{References}
\bibliographystyle{IEEEtran}
\bibliography{ref} 

\vspace{12pt}

\end{document}